\documentclass[letterpaper, 10 pt, conference, twoside]{IEEEtran}   

\IEEEoverridecommandlockouts                              

\usepackage{mathptmx} 
\usepackage{times} 
\usepackage{amsmath} 
\usepackage{amssymb}  

\usepackage{graphicx}

\usepackage{float}
\usepackage{multirow}
\usepackage{tikz}
\usepackage{cite}
\usepackage[acronym]{glossaries}
\usepackage{balance}
\usepackage{xcolor} 
\usepackage{soul}

\makeatletter
\let\NAT@parse\undefined
\makeatother
\usepackage[hyperfootnotes=false]{hyperref}

\title{\LARGE \bf
TSDF++: A Multi-Object Formulation for\\ Dynamic Object Tracking and Reconstruction
}

\author{Margarita Grinvald$^{1}$, Federico Tombari$^{2, 3}$, Roland Siegwart$^{1}$, and Juan Nieto$^{1}$
\thanks{This work was supported by a Google AR/VR University Research Award, ABB Corporate Research, and the European Union’s Horizon 2020 Research and Innovation Programme under Grant Agreement no. 101017008.}
\thanks{$^{1}$Margarita Grinvald, Roland Siegwart, and Juan Nieto are with the Autonomous Systems Lab, ETH Zurich, 8092 Zurich, Switzerland \mbox{(e-mail:~{\tt\small mgrinvald@ethz.ch}).}}%
\thanks{$^{2}$Federico Tombari is with the Technical University of Munich, Germany.}%
\thanks{$^{3}$Federico Tombari is with Google, Switzerland.}%
}

\newcommand{\acronym}[1]{\gls*{#1}\@}

\newacronym{tsdf}{TSDF}{Truncated Signed Distance Function}
\newacronym{icp}{ICP}{Iterative Closest Point}

\begin{document}


\maketitle

\IEEEpubid{\begin{minipage}{\textwidth}\ \\[10pt] \centering
  \\\copyright~2021 IEEE. Personal use of this material is permitted. Permission from IEEE must be obtained for all other uses, in any current\\ or future media, including reprinting/republishing this material for advertising or promotional purposes, creating new collective\\ works, for resale or redistribution to servers or lists, or reuse of any copyrighted component of this work in other works.
\end{minipage}}


\begin{abstract}

%
%
The ability to simultaneously track and reconstruct multiple objects moving in the scene is of the utmost importance for robotic tasks such as autonomous navigation and interaction.
Virtually all of the previous attempts to map multiple dynamic objects have evolved to store individual objects in separate reconstruction volumes and track the relative pose between them.
%
While simple and intuitive, such formulation does not scale well with respect to the number of objects in the scene and introduces the need for an explicit occlusion handling strategy.
In contrast, we propose a map representation that allows maintaining a single volume for the entire scene and all the objects therein.
%
To this end, we introduce a novel multi-object TSDF formulation that can encode multiple object surfaces at any given location in the map.
%
%
In a multiple dynamic object tracking and reconstruction scenario, our representation allows maintaining accurate reconstruction of surfaces even while they become temporarily occluded by other objects moving in their proximity.
We evaluate the proposed TSDF++ formulation on a public synthetic dataset and demonstrate its ability to preserve reconstructions of occluded surfaces when compared to the standard TSDF map representation.
%
%
%
Code is available at \url{https://github.com/ethz-asl/tsdf-plusplus}.

\end{abstract}
\IEEEpubidadjcol

\section{Introduction}
The task of reconstructing a dense map of the environment from a stream of RGB-D frames is key in the context of robotics, with applications in localization, autonomous navigation, interaction and manipulation. 
%
%
The widespread availability of low-cost RGB-D sensors and the advent of real-time methods for combining depth information into dense 3D models~\cite{newcombe2011kinectfusion} have together contributed to the rapid progress in the field in the last decade.

Most dense mapping systems, however, operate on the assumption of a static environment, that is, one in which the sensor is the only moving object. 
Dynamic parts of the scene are often regarded as outliers for the static model and are, if anything, purposely excluded from the mapping step. 
Naturally, such an approach does not hold for the highly-dynamic real-world environments that robots share with humans or other autonomous agents. 
Most notably, the method fails to discover and reconstruct elements that are often of highest relevance for navigation and interaction tasks, that is, the objects that are moving in the foreground and hence breaking the static assumption. 
%

\begin{figure}[t]
  \centering\includegraphics[width=0.97\columnwidth]{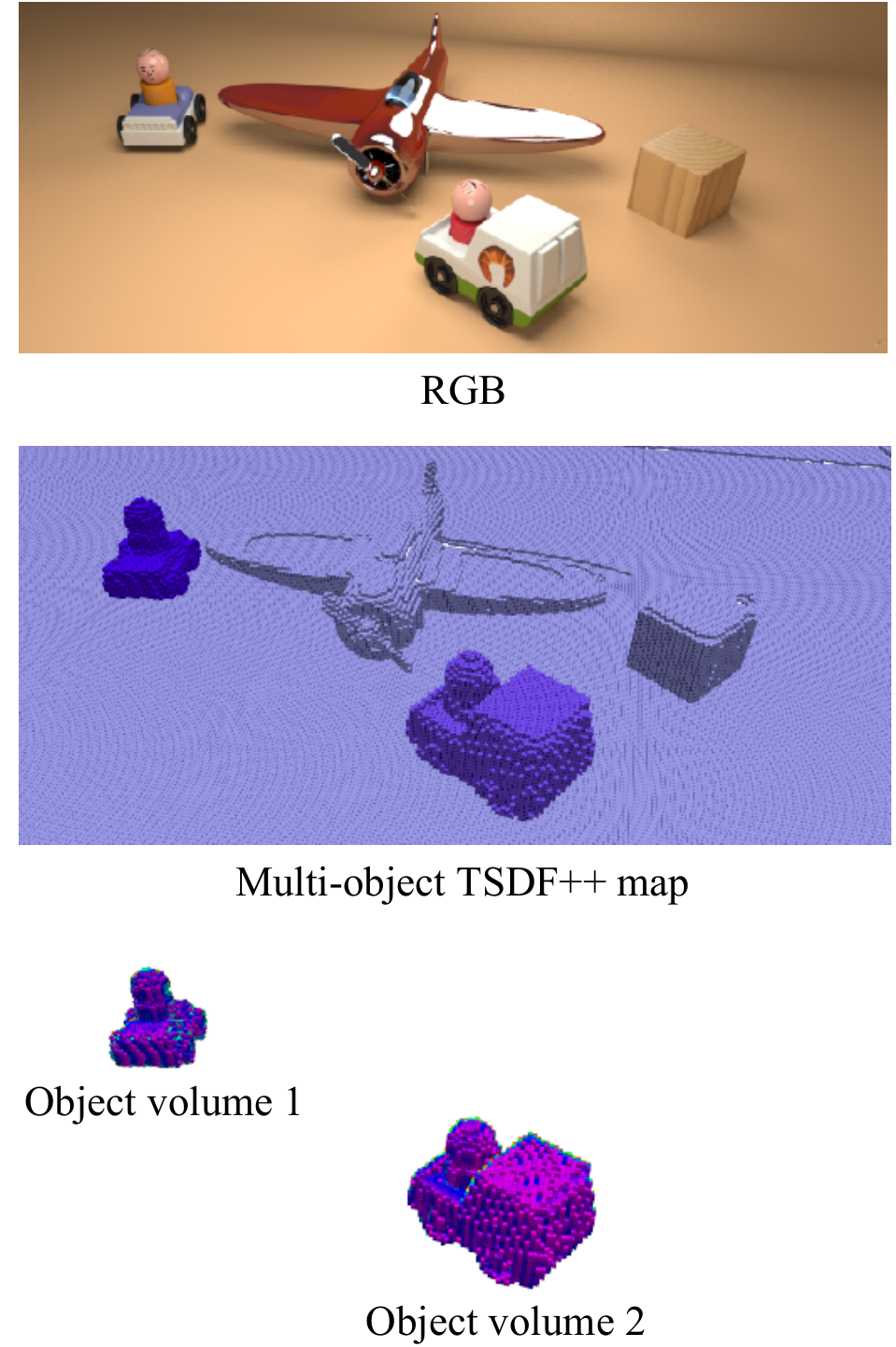}
  \caption{The proposed multi-object TSDF++ map representation is shown as a voxel grid in which voxels are colored based on the object reconstruction they index. The formulation allows storing multiple implicit object surfaces at each voxel and can be intuitively interpreted as a global map volume that indexes multiple smaller per-object TSDF volumes (TSDF values shown as color gradients) in a global coordinate frame. Progressive mapping results for the synthetic \textit{ToyCar3} sequence from~\cite{ruenz2017cofusion} are shown in the accompanying video available at \url{https://youtu.be/dSJmoeVasI0}.  }
  \vspace{-0.5cm}
\end{figure}

\IEEEpubidadjcol

To address this limitation, a number of works in the field of multiple rigid object tracking and reconstruction have recently emerged~\cite{ruenz2017cofusion, ruenz2018maskfusion, xu2019midfusion, strecke2019emfusion, barsan2018robust, miksik2019live}. 
%
%
%
They proposed a new paradigm for tracking the motion of a number of arbitrary objects moving independently in front of the camera and simultaneously densely reconstructing their 3D shape. 
%
Typically, the geometry of each individual object instance is reconstructed as a separate surfel cloud~\cite{ruenz2017cofusion, ruenz2018maskfusion} or \acronym{tsdf} volume~\cite{xu2019midfusion, strecke2019emfusion, barsan2018robust, miksik2019live}. 
%
%
The map of the entire scene, therefore, is stored as a collection of multiple 3D models and their associated poses with respect to some global reference frame.
As a result, updating the 3D map to reflect the latest estimated position of an object is as simple as updating the corresponding associated rigid transformation.

While simple and intuitive, such map representation can carry several limitations. 
For instance, each of the computationally expensive ray casting operations commonly required by the tracking-fusion pipeline must be performed multiple times, once for every object model stored in the map, thus scaling poorly with respect to the total number of objects in the scene. 
Moreover, ray casting through multiple independent volumes requires an explicit occlusion handling strategy to determine surface visibility when an object is occluded by another.
%
%
%
%

%

%

In this work, we introduce \acronym{tsdf}++ -- a novel map representation for the dynamic object tracking and reconstruction task that allows reconstructing the entire environment and the objects therein within a single 3D volume.
%
%
Ours is, to the best of our knowledge, the first multi-object \acronym{tsdf} formulation that can simultaneously store multiple implicit object surfaces at each voxel in the volume.
In particular, we aim to answer the following question: does a multi-object \acronym{tsdf} representation enable accurate surface reconstructions that can withstand occlusion in a multiple dynamic object tracking and reconstruction scenario?
%
%
This is contrast to a traditional \acronym{tsdf} map, in which surfaces that are temporarily occluded by other objects moving in their close vicinity become inevitably corrupted due to incoming conflicting observations of the occluding object geometry.
Further, by storing the entire scene within a single volumetric grid our mapping paradigm avoids the need to raycast several volumes at each time step and does not require an explicit occlusion handling strategy to determine which of the objects in the map are visible from the current camera view.
%

%
The proposed map formulation is experimentally evaluated in the context of a multiple dynamic object tracking and reconstruction task on the public synthetic dataset published with~\cite{ruenz2017cofusion}.
The results demonstrate the benefits of the proposed multi-object \acronym{tsdf}++ paradigm for densely reconstructing scenes containing multiple moving objects by validating its ability to maintain accurate reconstructions of surfaces even while they become occluded by another object.

To summarize, the main contributions of this work are:
\begin{itemize}
    \item A multi-object \acronym{tsdf} formulation encoding multiple object surfaces within a single 3D volume.
    \item A real-time CPU-only multiple dynamic object tracking and volumetric reconstruction framework.
    \item A novel map update strategy based on the proposed \acronym{tsdf}++ representation.
    \end{itemize}

\section{Related Work}
The last decade has seen considerable progress in the field of dense mapping, starting from the advent of efficient online depth fusion techniques.
%
%
%
%
KinectFusion~\cite{newcombe2011kinectfusion} was the first such framework to demonstrate the use of a commodity \mbox{RGB-D} sensor to incrementally reconstruct the 3D geometry of an indoor environment in real-time on a GPU.
%
%
%
The system produced compelling high-quality reconstructions by relying on an implicit \acronym{tsdf}-based volumetric representation~\cite{curless1996volumetric}.  
In general, \acronym{tsdf} maps allow for efficient incremental updates while systematically regularizing sensor measurement noise. 
Further, such representation is well-suited for robotic applications as it directly offers information about free space and surface connectivity, highly relevant for safe planning of navigation and interaction tasks.

The next big leap in the field was marked by the inclusion of rich semantic information into the reconstructed maps, largely enabled by recent advances in deep learning and GPU computing.
%
The first works proposed to incrementally fuse predictions from a 2D semantic segmentation network into a surfel-based representation~\cite{mccormac2017semanticfusion} or volumetric map~\cite{pham2019realtime} to obtain a dense semantically annotated reconstruction.
%
Subsequently, an even higher level of scene understanding has been pursued by introducing awareness of the individual object instances in the scene. 
The work in~\cite{sunderhauf17meaningful} relied on the SSD bounding box object detector~\cite{liu2016ssd} to build the first point-based object-oriented, online semantic mapping system.
Frameworks in~\cite{mccormac2018fusion} and~\cite{grinvald2019volumetric}, instead, make use of instance-aware semantic predictions from the Mask R-CNN network~\cite{he2017maskrcnn} to incrementally reconstruct a volumetric map of the environment at the level of individual objects.
Still, at the very core of all of these methods lies the key assumption of a largely static environment. 


The pioneering work in~\cite{newcombe2015dynamicfusion} represented an important step towards dense incremental reconstruction of scenes in which the assumption of a static environment is lifted. 
The framework demonstrated template-free real-time capturing of general non-rigid scene geometry with a single RGB-D sensor.
This prompted a flurry of follow-up research aimed at robustifying the method~\cite{innmann2016volumedeform, slavcheva2017killingfusion, slavcheva2018sobolevfusion} and extending it from a single camera to multiple views~\cite{dou2016fusion4d}.
The main focus of these works, however, are scenes containing a single deformable body moving in a non-rigid manner.

Recently, increasing attention has been given to densely reconstructing scenes containing multiple rigid objects of arbitrary shape which are independently moving in front of the camera.
%
Co-Fusion~\cite{ruenz2017cofusion} extends the surfel-based mapping framework in~\cite{whelan2016elasticfusion} to separately track and reconstruct objects segmented on a per-frame basis via geometric and motion cues. 
Similarly, MaskFusion~\cite{ruenz2018maskfusion} tracks and reconstructs multiple objects as separate surfel clouds, but the segmentation step combines a geometric approach with semantic pixel-wise object masks from \mbox{Mask R-CNN~\cite{he2017maskrcnn}}.
%
\mbox{MID-Fusion~\cite{xu2019midfusion}} adopts a similar strategy, but stores the tracked object models in \acronym{tsdf} volumes that have the advantage of being directly applicable for robot task planning. 
EM-Fusion~\cite{strecke2019emfusion} also opts for a \acronym{tsdf}-based representation to model the object geometries, but additionally offers a probabilistic data association scheme.
%
The works in~\cite{barsan2018robust} and~\cite{miksik2019live} extend the dynamic object reconstruction paradigm to larger outdoor scenes.

While offering a diverse set of approaches to the task of multiple object tracking and reconstruction, all of these frameworks share a common map representation paradigm. 
%
%
They store individual object instances in separate and independent reconstruction volumes, be those surfel clouds~\cite{ruenz2017cofusion, ruenz2018maskfusion} or \acronym{tsdf} grids~\cite{xu2019midfusion, strecke2019emfusion, barsan2018robust, miksik2019live}.
The resulting environment map is thus simply a collection of multiple volumes with their associated poses: one for the static background, and one for each of the detected objects in the scene.
Such object-centric representation undoubtedly offers several advantages.
It is relatively intuitive, and allows reconstructing each object at the most suitable level of detail by choosing size-dependent resolutions for each volume.
Further, to update the pose of a tracked object in the map it is enough to change the rigid transformation associated to its corresponding 3D model.

At the same time, the paradigm of reconstructing each object as a separate volume can carry several limitations. 
For every incoming frame, the tracking, data association, and visualization steps of a typical dynamic object reconstruction pipeline each require one ray casting operation to determine the visibility of the surfaces stored in the map.
With the map consisting of several independent volumes, every ray casting operation translates to repeatedly casting the same rays through all of them.
This includes volumes that are not visible from the given camera view, since it is not known \mbox{\textit{a priori}} which of the objects are in view and which are not.
%
%
Related to this is the need for an explicit occlusion handling strategy to determine whether a volume is, fully or partially, occluded by another such that the visibility of its surfaces can be properly resolved.

In contrast to all existing methods for multiple dynamic object tracking and reconstruction, our approach proposes to reconstruct all of the tracked objects in a single volume, thus removing the need to repeatedly ray cast through each of the mapped models multiple times and eliminating entirely the occlusion handling problem.
This is made possible with a novel multi-object \acronym{tsdf} formulation that allows storing multiple object surfaces at each of the voxels in the 3D grid.
To the best of our knowledge, ours is the first dense mapping approach that can simultaneously maintain the implicit representation of multiple object surfaces that are in contact or close vicinity within a single volume.
This novel map representation is highly relevant for reconstructing multiple moving objects, as it enables preserving reconstructions of surfaces even while they are temporarily occluded by another object.

\begin{figure}[t]
  \centering\includegraphics[width=\columnwidth]{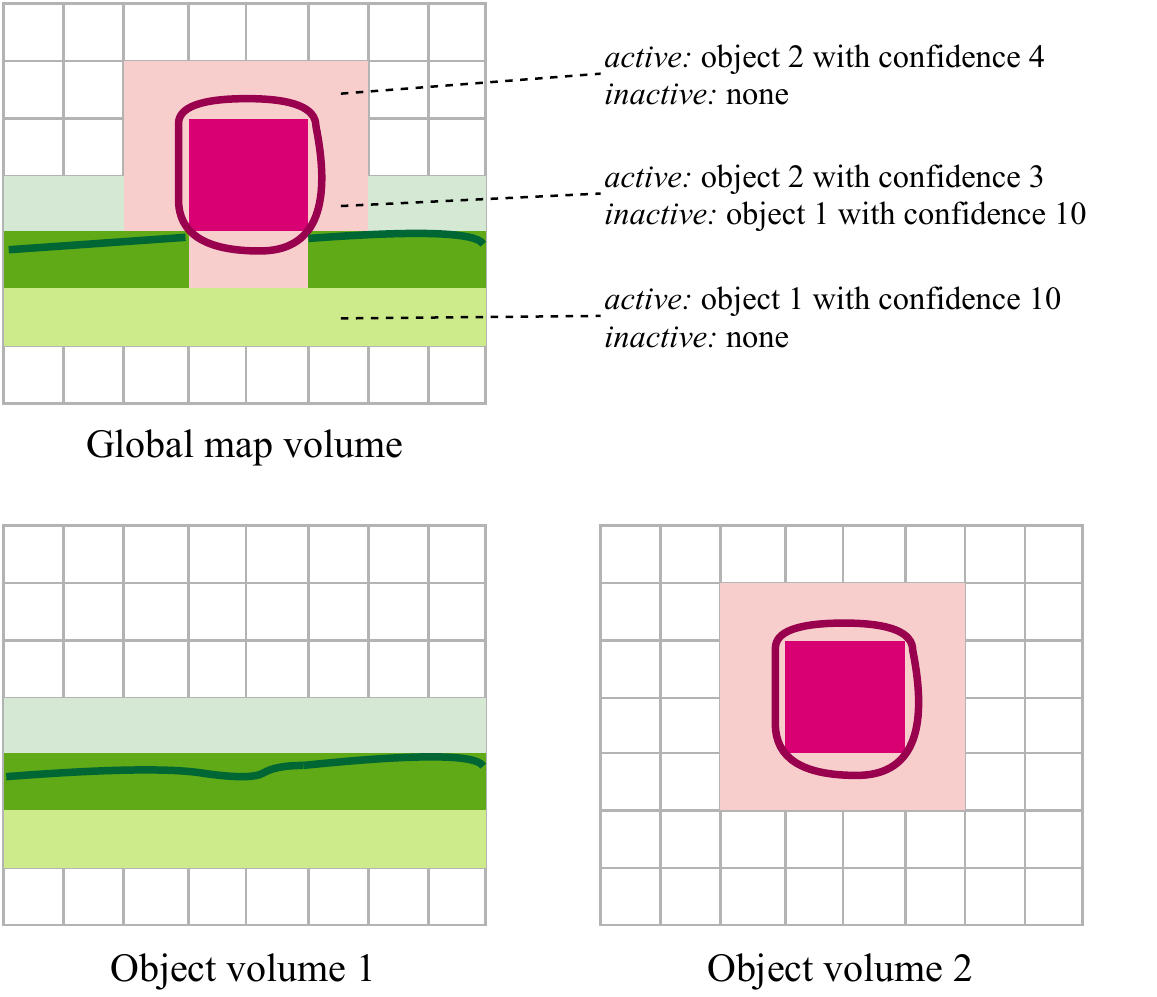}
  \caption{The proposed \acronym{tsdf}++ formulation can be visualized with a toy example in 2D. The representation consists of several object volumes, one for each mapped object, and a global map volume that indexes all of them within a common global coordinate frame. Each voxel in the global map volume keeps track of which object model is currently \textit{active} and which object model, if any, is \textit{inactive} or temporarily occluded at that location. By storing the \acronym{tsdf} reconstruction (here shown via color gradients) of objects across separate volumes, the formulation is able to maintain reconstruction of surfaces in the map even if they become temporarily occluded by another object moving in their near proximity. }
  \label{figure:tsdf-plusplus}
\end{figure}

\section{Method}

The proposed framework for multiple dynamic object tracking and reconstruction consists of the novel \acronym{tsdf}++ map representation and a pipeline to incrementally process the RGB-D stream of a localized\footnote{Note that the current work focuses on the mapping aspect of the pipeline, hence localization of the camera is assumed to be given.} depth camera. 
%
%
%

The pipeline deployed at each incoming frame consists of four steps: (i)~per-frame object segmentation, (ii)~data association, (iii)~object pose tracking, and (iv)~map update. 
%
First, the incoming RGB-D image pair is segmented via a combined geometric-semantic scheme to identify individual object instances.
%
%
Next, the detected object segments are matched to the object models stored in the map via a data association strategy.
Given the correspondences between the mapped object models and their observations in the incoming frame, the rigid pose of each object is tracked via \acronym{icp}.
Finally, the map is updated to reflect the latest estimated pose of each object instance and the depth and segmentation observations are fused into the volume.
%
%



%

\subsection{TSDF++ formulation}
The traditional \acronym{tsdf}-based map representation consists of a voxel grid in which each voxel stores the projective signed distance to the closest observed surface, truncated up to a threshold.
The sign denotes whether the voxel center lies behind or in front of the surface.s
Previous works, like the one in~\cite{grinvald2019volumetric}, have augmented the representation to additionally store at each voxel information about which object instance it belongs to, demonstrating incremental reconstruction of multiple static objects within a single \acronym{tsdf} volume.
In general, the same idea can be directly applied to a multiple dynamic object tracking and reconstruction scenario, by simply translating within the volume the \acronym{tsdf} values associated with a moving object to reflect its latest tracked pose.
However, a key limitation of the traditional \acronym{tsdf} representation with this approach is that the translated values would effectively overwrite whatever surface reconstruction was previously stored at the destination location.
To intuitively illustrate this with an example, this means that a large portion of a tabletop surface that has been densely reconstructed over time would become gradually corrupted and eventually be erased if an object, such as a laptop, would move and occlude it.
Once the laptop is removed, the corresponding portion of the table surface would need to be reconstructed once again, entirely from scratch.

To overcome this limitation we propose \acronym{tsdf}++, a novel \acronym{tsdf} formulation that allows storing at each voxel more than one observed object surface.
In the context of a multiple dynamic object tracking and reconstruction task, this means that we can store at each voxel the \acronym{tsdf} value of the object model that is visible at the corresponding location in the map, as well as the \acronym{tsdf} value of other objects that are currently occluded by the first one.
Intuitively, our novel multi-object \acronym{tsdf}++ formulation can be interpreted as a layered volumetric representation, with the first layer encoding the visible object surfaces and the remaining layers storing the occluded surfaces.
In this work, we limit the number of object surfaces stored at each voxel to two, as this is sufficient for most practical applications.
At the same time, the representation can be easily extended to store more layers in order to cope with higher degrees of scene clutter.

In practice, the \acronym{tsdf}++ representation is implemented as a collection of multiple smaller \acronym{tsdf} volumes, one for each mapped object model, which are all indexed via a global scene-size volume, referred to from now on as the global map volume.
Each voxel in the global map volume keeps track of which object model is \textit{active} at the corresponding 3D location and which object model, if any, is \textit{inactive} or occluded.
Additionally, for both the \textit{active} and the \textit{inactive} object, the voxel stores an associated integer confidence score.
The individual object volumes encode the actual \acronym{tsdf} of the corresponding object geometry in a common global coordinate frame and share the same resolution as the global map volume.
As a result, given a voxel in the global map volume, the corresponding \acronym{tsdf} voxel in one of the object volumes can be directly accessed using the exact same grid coordinates.
An illustration of this formulation can be found in Figure \ref{figure:tsdf-plusplus}.
%
Both the \acronym{tsdf}++ representation and the corresponding fusion strategy are implemented as extensions of the \acronym{tsdf}-based voxel hashing approach in~\cite{oleynikova2017voxblox}.

%
%
%

\subsection{Per-frame geometric-semantic segmentation}
\label{section:segmentation}
The goal of the per-frame segmentation step is to identify individual object instances in an RGB-D image pair and extract the corresponding segments. 
In a multiple dynamic object tracking and reconstruction scenario the extracted segments are later used to track the pose of the corresponding object models and to update their 3D reconstruction.
As a result, even slight inaccuracies in the per-frame segmentation results exhibit a  compounding effect, significantly impacting the tracking and reconstruction performance in all subsequent frames.
For this reason, it is of the uttermost importance for the extracted segments to accurately reflect the true contours of the objects in the scene.
To this end, the same combined geometric-semantic scheme as the one presented in~\cite{grinvald2019volumetric} is deployed at each incoming frame, processing the color and the depth images in parallel.

The RGB frame is processed with the Mask R-CNN~\cite{he2017maskrcnn} instance-aware semantic segmentation network to predict for each object instance a semantically annotated pixelwise mask.
While the learning-based segmentation approach is able to deal with large intra-class variability, the predicted segmentation masks are known to suffer from inaccurate contours.
Namely, the masks either bleed into the background or undershoot with respect to the true outline of an object.
%

At the same time, the corresponding depth frame is processed with the purely geometric normal-based approach from \cite{furrer2018incremental} to extract a set of convex object-like segments.
%
%
%
%
The resulting segments accurately reflect the true contours of the corresponding real-world objects, but often exhibit over-segmentation of non-convex articulated object categories.

In order to combine the complementary nature of the semantically-annotated object masks predicted with \mbox{Mask R-CNN} and the geometrically-accurate convex segments, the two segmentation outputs are fused into one.
%
%
The pairwise 2D overlap between the predicted object instance masks and the extracted segment clouds is computed, normalized by the number of points in the segment cloud.
Whenever such overlap is higher than a threshold $\tau _\mathrm{overlap}$, which we set at $80\%$, the segmented cloud is matched to the corresponding object instance, and finally all segments matching to a same instance are merged into one.
%
The result is a set of segment point clouds which accurately reflect the true shape of the observed scene objects, with some of them having an associated semantic category.

\subsection{Data association}
Because the per-frame segmentation strategy processes each incoming RGB-D frame independently, it fails to provide an association between the corresponding segments and object instances predicted across the different frames. 
To track segments and instances across frames we adopt a data association strategy similar to the one presented in~\cite{grinvald2019volumetric}, which relies on the known camera pose to lookup the spatially corresponding voxel for each segment point and employs a voting scheme to match the input segments to one of the mapped object instances.
A key difference to the approach in~\cite{grinvald2019volumetric} is that in this work we allow for multiple incoming 3D segments to match to a same mapped object instance.
Further, all segments in the current frame that map to the same object instance are merged together into a single segment.
This is intended to alleviate the negative impact on the tracking and reconstruction accuracy of those frames for which \mbox{Mask R-CNN} fails to predict a segmentation mask for articulated, non-convex objects, which thus remain over-segmented by the geometric scheme.
%


\subsection{Object model pose tracking}
We track the pose of all the objects which are visible in the current frame by registering the extracted 3D segments to their corresponding object models in the map.
While the proposed frame-wise geometric semantic segmentation scheme would allow to track all the segmented parts of a scene, to limit this step to the foreground objects we only track the pose of those object models which have been, at some point, detected and semantically classified.

Given an input segment and its corresponding object model, we register the 3D points of the segment expressed in the global frame to the vertices of the mesh extracted from the \acronym{tsdf} volume of the mapped object to compute the rigid transformation $\mathbf{T}_\mathrm{O}$ that represents the relative transformation of the object model pose between the previous and the current frame.
We rely on the ICP implementation from the Point Cloud Library~\cite{holz2015registration} and adopt the point-to-plane error metric, which can be expressed as:
\begin{align}
    E(\mathbf{T}_\mathrm{O}) = \sum_{k=1}^N ((\mathbf{T}_\mathrm{O}\, \mathbf{s}_k - \mathbf{o}_k) \cdot \mathbf{n}_{\mathbf{o}_{k}})^2
\end{align}
where $\mathbf{s}_k$ is a point in the 3D segment, and $\mathbf{o}_k$ and $\mathbf{n}_{\mathbf{o}_{k}}$ are, respectively, the corresponding vertex and local surface normal in the mesh extracted from the object model \acronym{tsdf}. 
$N$ is the total number of points in the segment point cloud for which a corresponding point could be identified in the target set of mesh vertices.
The error term is minimized with the Levenberg-Marquardt non-linear optimizer. 

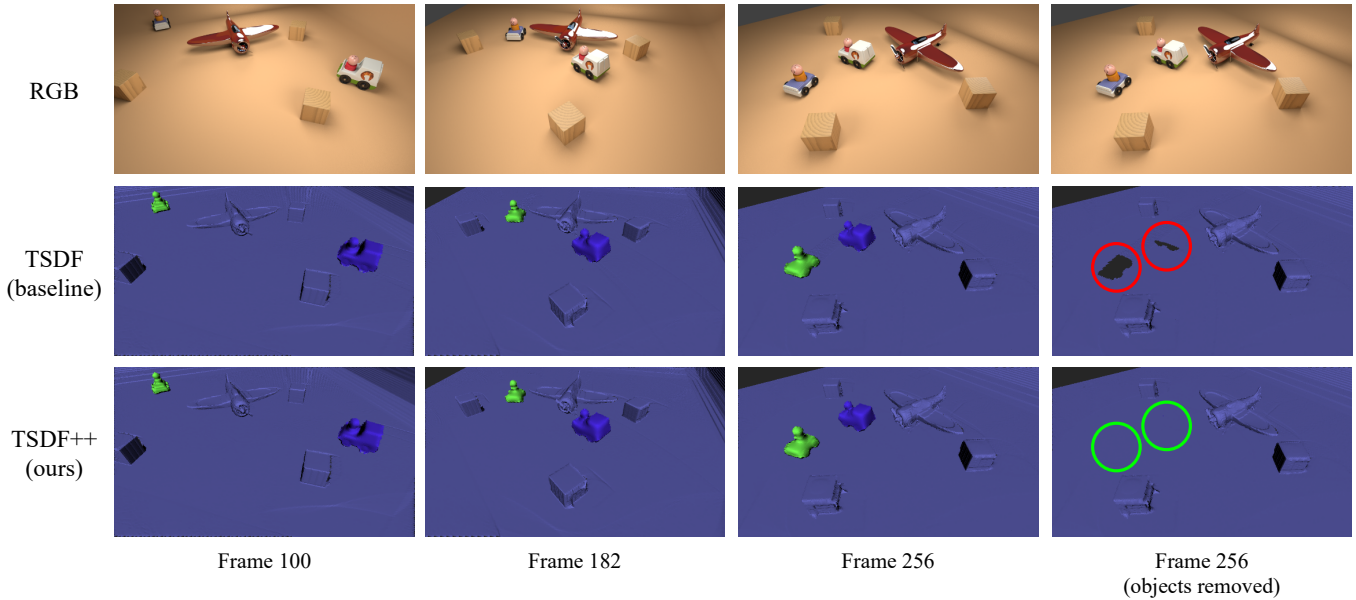
\begin{figure*}
\begin{tikzpicture}[every node/.style={inner sep=0,outer sep=0}]
\node(c){\includegraphics[width=0.99\textwidth]{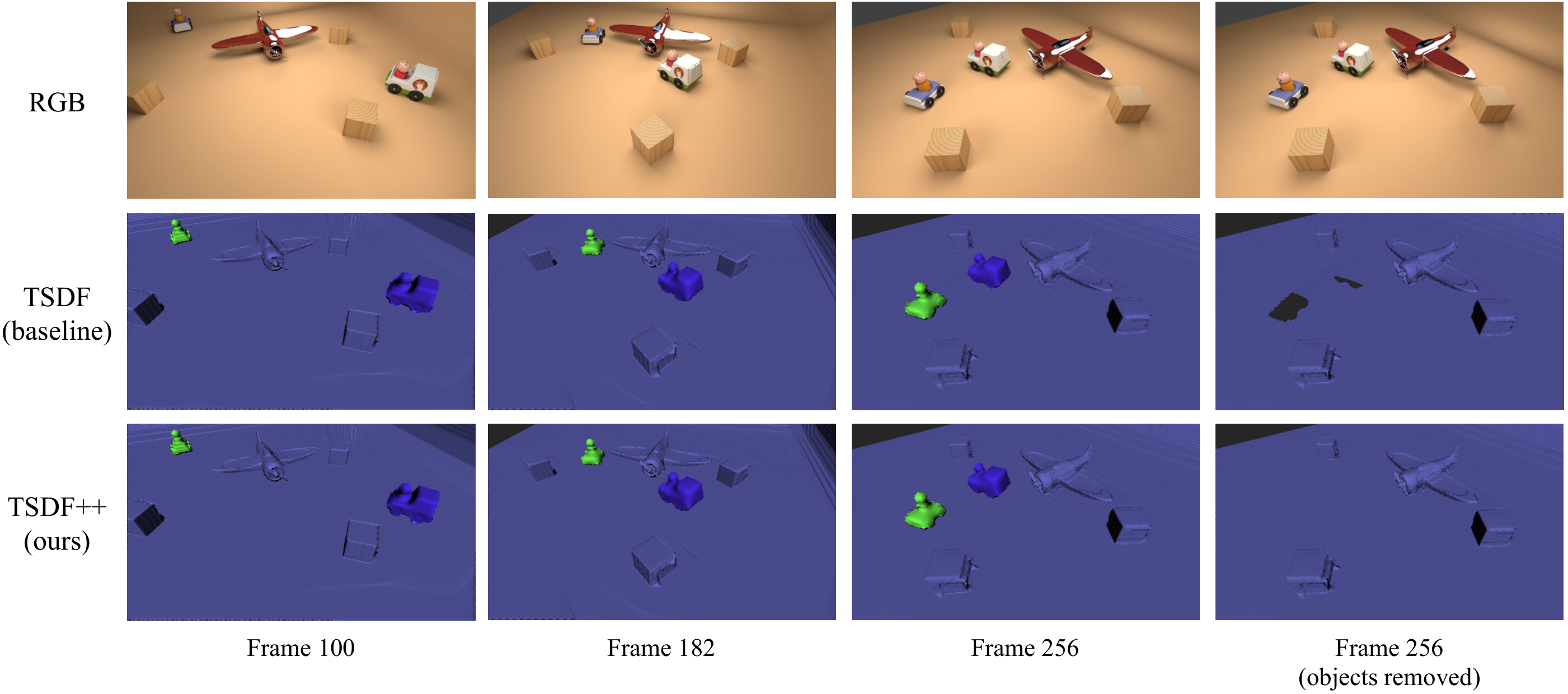}};
\node at(c.center) [
    circle,
    draw,
    red,
    very thick,
    minimum size=4ex,
    yshift=13pt,
    xshift=165pt
] {};
\node at(c.center) [
    circle,
    draw,
    red,
    very thick,
    minimum size=4ex,
    yshift=21pt,
    xshift=184pt
] {};
\node at(c.center) [
    circle,
    draw,
    green,
    very thick,
    minimum size=4ex,
    yshift=-55pt,
    xshift=165pt
] {};
\node at(c.center) [
    circle,
    draw,
    green,
    very thick,
    minimum size=4ex,
    yshift=-47pt,
    xshift=184pt
] {};
\end{tikzpicture}
\caption{The proposed \acronym{tsdf}++ map representation is compared against the standard \acronym{tsdf} formulation within the presented dynamic object tracking and reconstruction framework. The progressive mapping results on the synthetic \textit{ToyCar3} sequence from \cite{ruenz2017cofusion} are shown in the form of a mesh extracted from the volumetric representation, with different colors representing individual object instances. We rely on the ground truth segmentation provided with the sequence and reconstruct the scene up to frame 256. Next, we simulate the removal of the two moving toy car objects to reveal the underlying tabletop surface and verify whether its reconstruction has been preserved throughout occlusion. The standard \acronym{tsdf} representation presents holes in the surface of the table despite those parts having been previously reconstructed, thus requiring those portions of the surface to be reconstructed from scratch in the subsequent frames.
In contrast, the proposed \acronym{tsdf}++ formulation reveals an intact reconstruction of the tabletop surface, thanks to its ability to maintain accurate surface reconstructions even while these are temporarily occluded by other objects moving in the near proximity.}
\label{figure:experiment}
\vspace{-0.01cm}
\end{figure*}

\subsection{Object model pose update}
Once the relative rigid transformation $\mathbf{T}_\mathrm{O}$ between the pose of an object model at the previous and the current frame has been estimated, the map of the scene needs to be updated accordingly.
Since the proposed \acronym{tsdf}++ representation stores all of the mapped object models within a same grid and coordinate frame, updating the pose of an object requires actually translating the corresponding \acronym{tsdf} values within the tracked object volume and updating the global map volume to reflect this change.

First, the object model whose pose is being updated is \textit{deactivated} from all of the voxels in the global map volume.
That is, in all the global map volume voxels which listed the translated object model as the \textit{active} one, the object model reference is removed and the model that was referenced as \textit{inactive}, if any, is \textit{activated}.
Next, the \acronym{tsdf} values in the corresponding object model volume are translated according to the estimated relative rigid transformation $\mathbf{T}_\mathrm{O}$.
When translating \acronym{tsdf} values from a source location to a destination location within a discretized grid, there is no guarantee of a one-to-one mapping between source and target voxel centers. 
Instead, computing the \acronym{tsdf} value at a destination voxel center can be achieved through interpolation of the \acronym{tsdf} values around the corresponding source location.
We adopt the trilinear interpolation method to compute the \acronym{tsdf} values at those destination voxel centers for which the corresponding source location offers eight valid neighbouring voxels required by the computation.
%
Finally, the global map voxels in correspondence of the destination location are all updated to \textit{deactivate} the previously \textit{active} object model and instead \textit{activate} the translated one.

%

\subsection{Map fusion}
Finally, the 3D segments extracted from the input \mbox{RGB-D} frame are fused into the \acronym{tsdf}++ map.
After projecting the segments into the global volume map using the known camera pose, voxels corresponding to each projected 3D point are updated to store the incoming segmentation and depth information.
%
%
First, the corresponding global map volume voxel is updated to reflect the observed segmentation information.
If the ID of the incoming object segment agrees with the ID of the currently \textit{active} object model the corresponding confidence score is increased by one unit, else it is decreased by one unit.
If the confidence reaches zero, the object volume corresponding to the incoming object ID becomes the new \textit{active} object volume at that voxel, and the corresponding confidence score is re-initialized to one.
Next, the volume corresponding to the \textit{active} object at that voxel is retrieved or newly allocated.
The \acronym{tsdf} value and weight for the voxel at the corresponding location in the \textit{active} object volume is finally updated using the standard weighted average scheme.

%


\section{Experiments}
The proposed framework for multiple dynamic object tracking and reconstruction, together with the novel \acronym{tsdf}++ map representation, is evaluated on the synthetic dataset published with~\cite{ruenz2017cofusion}.
The system is evaluated on a laptop with an Intel Xeon E3-1505M eight-core CPU at 3.00\,GHz with 16\,GB memory.
The Mask R-CNN inference code runs on an Nvidia Quadro M2200 GPU with 4\,GB of memory and is based on the publicly available implementation from Matterport\footnote{\href{https://github.com/matterport/Mask_RCNN}{https://github.com/matterport/Mask\_RCNN}} and the pre-trained weights provided for the Microsoft COCO dataset~\cite{lin2014coco}.
%
%
In our experiment the voxel size has been set to 1\,cm and the truncation distance to ten times the voxel size.

%



In the context of a multiple dynamic object tracking and reconstruction task, a key benefit of the proposed \acronym{tsdf}++ representation is the ability to maintain accurate volumetric reconstructions of surfaces even if they become temporarily occluded by other objects moving in the near proximity.
Once the occluding object moves away, the revealed surface is simply re-activated and thus does not need to be reconstructed from scratch.
In contrast, the traditional \acronym{tsdf} map representation applied to a dynamic object scenario inevitably leads to occluded surfaces being destroyed by the translated reconstruction of the moving occluding object.

In this regard, we evaluate the ability of the proposed map representation to preserve a previously reconstructed surface during occlusion and compare against the standard \acronym{tsdf}-based formulation.
In our experiments, we seek to maximally isolate the contribution of each of the two map representations from any noise that could be introduced by the per-frame segmentation scheme.
To this end, we adopt the synthetic \textit{ToyCar3} sequence published by the authors of Co-Fusion~\cite{ruenz2017cofusion} and rely on the accompanying ground truth per-frame segmentation.

%
%
%

The sequence of RGB-D frames is processed with our multiple object tracking and reconstruction framework up to frame 256.
The progressive object-level mapping results are illustrated in Figure~\ref{figure:experiment}.
Next, we simulate the removal of the two reconstructed toy car objects to reveal the tabletop surface underneath and verify whether it is still accurately reconstructed.
In Figure~\ref{figure:experiment}, it can be seen that the standard \acronym{tsdf}-based map representation reveals two holes in the table surface in correspondence of the two objects, despite those parts having been previously accurately reconstructed.
In contrast, the proposed \acronym{tsdf}++ formulation successfully maintains a complete and accurate reconstruction of the tabletop surface even while it is occluded by other objects moving on top of it.
The ability of our multi-object volumetric map representation to preserve reconstructed surfaces throughout occlusion allows us to benefit from all the previously fused depth measurements whenever the occluded surfaces become visible again, rather than having to re-build their implict representations from scratch.
%


%
%
%
%

Additionally, we show qualitative results for incremental tracking and reconstruction on the entire \textit{ToyCar3} sequence from~\cite{ruenz2017cofusion} with the full proposed pipeline (including the per-frame segmentation scheme) in the accompanying video.
The framework is able to process the sequence of RGB-D frames of resolution 960x540 at a frequency of approximately 1\,Hz, which is comparable to the runtime performance of the framework for reconstruction of multiple static objects presented in~\cite{grinvald2019volumetric}.
It can be observed that the reconstruction results are heavily impacted by the inaccuracies introduced by the per-frame segmentation scheme, which in turn decreases the tracking performance.
In general, the ability to reconstruct multiple implicit object surfaces at any given location in the map comes at the cost of decreased robustness towards wrong or missing per-frame object segmentations.

\section{Conclusion}


This work presents \acronym{tsdf}++, a novel multi-object volumetric map representation that allows maintaining multiple implicit object surfaces at any given location in the map.
In contrast to previous approaches for mapping multiple objects moving in front of the camera, we propose to store all reconstructed object models within a single volume, thus removing the need to repeatedly ray cast through several object volumes at each timestep.
We demonstrate that the proposed formulation is particularly well suited for the multiple dynamic object tracking and reconstruction task in that it allows to preserve accurate reconstructions of surfaces even if they become temporarily occluded by other objects moving in their proximity.
%
%

The problem of simultaneously tracking and reconstructing multiple arbitrary objects moving in front of the camera is a multi-faceted one, requiring successful integration of several components, from per-frame object segmentation to camera pose tracking, from tracking of the pose of multiple objects to accurately reconstructing their 3D shape.
Ours is but a first step towards a complete pipeline and potential future work includes extending the proposed map representation and framework to increase robustness to inaccurate per-frame instance segmentation.
%
At the same time, we believe that accurate reconstructions of object surfaces that can withstand occlusion are key for unlocking unprecedented levels of scene understanding, and we hope that this work can pave the way towards exploration of novel representations for other object-level perception tasks.

%
%
%

\balance 


\bibliographystyle{IEEEtran}
\bibliography{literature}

\end{document}